\documentclass{article}

\usepackage[preprint]{style}

\usepackage[utf8]{inputenc} 
\usepackage[T1]{fontenc}    
\usepackage{hyperref}       
\usepackage{url}            
\usepackage{booktabs}       
\usepackage{amsfonts}       
\usepackage{nicefrac}       
\usepackage{microtype}      
\usepackage{xcolor}         
\usepackage{natbib}         
\usepackage{bm}

\usepackage{graphicx}
\graphicspath{ {figures/} }
\usepackage{amsmath,amssymb}
\DeclareMathOperator*{\argmax}{arg\,max}

\DeclareMathOperator*{\calL}{\mathcal{L}}
\DeclareMathOperator*{\rhovec}{\bm{\rho}}

\title{Bayesian Neural Network Versus Ex-Post Calibration For Prediction Uncertainty}

\author{%
  Satya Borgohain \\
  Monash University \\
  \texttt{satya.borgohain@monash.edu} \\
   \And
   Klaus Ackermann \\
   Monash University \\
   \texttt{klaus.ackermann@monash.edu} \\
   \AND
    Ruben Loaiza-Maya \\
    Monash University \\
   \texttt{ruben.loaizamaya@monash.edu} \\
}

\begin{document}

\maketitle

\begin{abstract}
Probabilistic predictions from neural networks which account for predictive uncertainty during classification is crucial in many real-world and high-impact decision making settings. However, in practice most datasets are trained on non-probabilistic neural networks which by default do not capture this inherent uncertainty. This well-known problem has led to the development of post-hoc calibration procedures, such as Platt scaling (logistic), isotonic and beta calibration, which transforms the scores into well calibrated empirical probabilities. A plausible alternative to the calibration approach is to use Bayesian neural networks, which directly models a predictive distribution. Although they have been applied to images and text datasets, they have seen limited adoption in the tabular and small data regime. In this paper, we demonstrate that Bayesian neural networks yields competitive performance when compared to calibrated neural networks and conduct experiments across a wide array of datasets.

\end{abstract}

\section{Introduction}\label{sec:intro}

Obtaining well calibrated estimates of probabilities is crucial, particularly in domains where decision making could have direct consequences on human life such as medical diagnosis \cite{Huang2020} and credit approval \cite{beque2017approaches} to name a few. In such scenarios, it is often not enough for a model to be accurate but it also needs to capture and quantify the degree of uncertainty with which it makes such decisions.  However, many classification tasks still primarily rely on the classifier's error rate (or accuracy) as the key metric for its selection and deployment in real world use-cases which could lead to over/under confident predictions. Coupled with class imbalance it poses challenges that given a set of features, simply predicting class membership does not help us understand. Raw probabilities estimates provide a more complete picture of the underlying decision making process by the classifiers and alternative diagnostic metrics such as AUC-ROC, precision, recall and F1 certainly help in that regard \cite{kuhn2013applied}. Post-hoc calibration methods are one such approach which helps match the predicted probabilities with the expected class distribution of the target. They take the output of any model and map the score to the empirical probabilities \cite{Kull2017}.

Fairness and bias is another closely related field in machine learning and an arena that has received much attention recently. Some of the methods predominantly used there involve thresholding in order to de-bias the learned model against certain classes. However, finding optimal thresholds without optimization are only possible if the classifiers are well-calibrated \cite{Kull2017}.

Bayesian neural networks (BNN) have recently emerged as an alternative approach to calibration methods \cite{kingma2013auto}. However, adding Bayesian layers to neural networks alone, does not solve the problem of not receiving well calibrated output as shown in \cite{hortua2020parameter}. A BNN creates a probabilistic model by linking the neural net to the conditional distribution of the outcome of interest. Thus, BNNs directly allow the researcher to measure aleatoric uncertainty without the need for any post-hoc processing steps. Because BNNs are probabilistic models with high-dimensional parameters spaces they are estimated using approximate Bayesian methods such as variational inference  \citep{kingma2013auto}. The computation of a posterior distribution implies that BNNs also have the ability to capture epistemic uncertainty. Therefore, unlike calibration methods, the coherent probabilistic  nature of BNNs implies that both aleatoric and epistemic uncertainty are considered in the production of out-of-sample predictions. Furthermore, methods such as SWA-Gaussian \cite{maddox2019simple} which tries to approximate the posterior over the parameters using information inherent in the trajectory of SGD have provably shown that Bayesian methods do work well with neural networks to provide well-calibrated probabilities. Platt scaling has also seen much adoption in large neural nets \cite{guo2017calibration}.

Despite the modelling benefits of BNNs, their uptake in machine learning has been relatively slow. One of the main reasons is that, to our knowledge, no comprehensive studies have been undertaken to demonstrate the effectiveness of BNNs over calibration approaches, particularly for tabular and smaller datasets. The main purpose of this paper is to fill this gap in the literature. Although there have been notable developments using Bayesian approaches in the field, they usually involve non-tabular datasets with a large sample size.

Using a subset of the rich amalgam of datasets considered in \cite{Kull2017}, we investigate if BNNs outperform post-hoc calibration methods in terms of probability calibration.

The key contributions of the paper can thus be summarized as follows:
\begin{itemize}
    \item We illustrate that BNNs are a plausible alternative to obtaining well-calibrated empirical probabilities for classification.
    \item We specifically demonstrate the applicability of BNNs in tabular, real-world, small data regime.
\end{itemize}

\section{Bayesian neural networks}

\begin{figure}
  \centering
    \includegraphics[width=\textwidth]{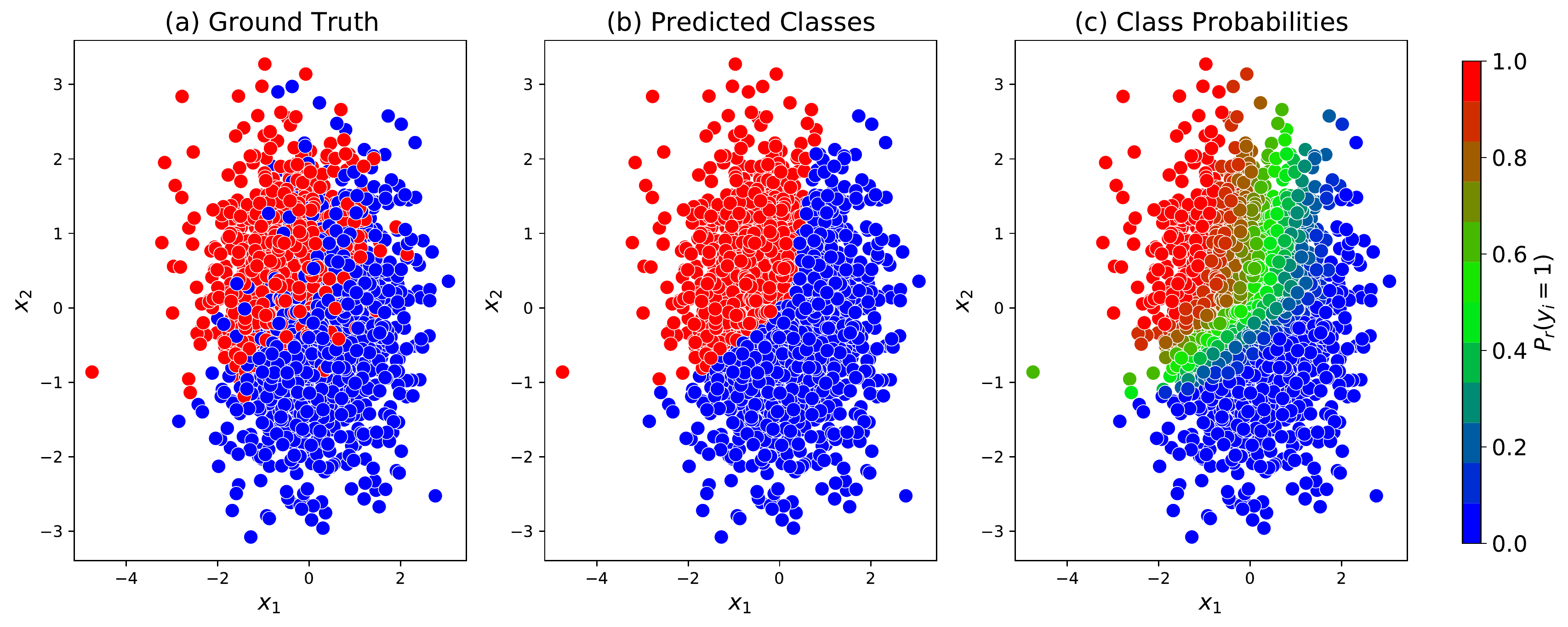}
  \caption{BNN predictive results for simulation example. Panel (a) displays the true classes, with red corresponding to observation with $y_i=1$ and blue to observations with $y_i=0$. Panel (b) presents the BNN point predictions $\hat{y}_i$, with the same color scheme as panel (a). Panel (c) reports the classification probabilities $\text{Pr}(Y_i=1|\bm{x}_i,\bm{\theta})$.}
  \label{Fig:toyexample}
\end{figure}

\subsection{The model}
\label{sec:model}
Denote as $\bm{y} = \left(y_1,\dots,y_N\right)^\top$ a vector of $N$ realizations of the binary variable $y_i\in\{0,1\}$, as $\bm{x}_i=\left(x_{1,i},\dots,x_{p,i}\right)^\top$ a vector of $p$ covariates with explanatory power on $y_i$, and set $\bm{x} = \left(\bm{x}_1^\top,\dots,\bm{x}_N^\top\right)^\top$. A Bayesian neural network (BNN) is a probabilistic model that links a neural network function $f_{NN}\left(\bm{x}_i,\bm{\theta}\right)$ to the conditional distribution function of $y_i$ (see \cite{mullachery2018bayesian} for an overview on BNNs). For the binary outcomes considered in this paper, we link $f_{NN}\left(\bm{x}_i,\bm{\theta}\right)$  to the conditional probability distribution of $y_i$ via the logistic function $g\left(a\right) = \frac{1}{1+e^{-a}}$, so that
\begin{equation}\label{EQ1}
	p\left(y_i|\bm{x}_i,\bm{\theta}\right) = g\left[f_{NN}\left(\bm{x}_i,\bm{\theta}\right)\right]^{I\left(y_i=1\right)}\left\{1-g\left[f_{NN}\left(\bm{x}_i,\bm{\theta}\right)\right]\right\}^{I\left(y_i=0\right)},
\end{equation} 
where $I\left(.\right)$ denotes an indicator function that is equal to one if its argument is true, and zero otherwise. Because a BNN fully characterizes the conditional distribution $p\left(y_i|\bm{x}_i,\bm{\theta}\right)$, it has the ability to both, produce point classification predictions as $\hat{y}_i = \argmax_{Y\in\{0,1\}} p(Y|\bm{\theta},\bm{x}_i)$, and also capture the level of classification uncertainty over those predictions as $\text{Pr}(\hat{y}_i = y_i|\bm{\theta},\bm{x}_i)$. 

Using the assumption that the elements in $\bm{y}$ are conditionally independent, we can then express the likelihood function for the probabilistic neural network as
\begin{align*}
p\left(\bm{y}|\bm{\theta},\bm{x}\right) =& \prod_{i=1}^{N}p\left(y_i|\bm{x}_i,\bm{\theta}\right).
\end{align*}
The main challenge in using a BNN is inference. The parameter vector $\bm{\theta}$ has generally thousands of elements, which often makes exact Bayesian estimation  infeasible. In the following section we discuss how approximate Bayesian estimation of BNNs can be applied instead.
\subsection{Variational inference}
In Bayesian estimation the density of interest is that of the parameters of the neural network conditional on the data. This density is denoted here by $p\left(\bm{\theta}|\bm{y}\right)\propto p\left(\bm{y}|\bm{\theta}\right)p\left(\bm{\theta}\right)$, where $p\left(\bm{\theta}\right)$ is the prior density and $\bm{x}$ is dropped for ease of notation . Because of the high-dimensionality of $\bm{\theta}$, exact Bayesian estimation methods are computationally costly, and as such not practical for the problem at hand.
Instead, we resort to variational inference (VI) methods, 
where a density $q_\lambda(\bm{\theta})$ - member of some parametric family of densities - is used to 
approximate $p(\bm{\theta}|\bm{y})$, and where $\bm{\lambda}$ is a vector of parameters  known as variational parameters (see for instance \cite{blei2017variational} and \cite{kingma2013auto}).
Variational inference can then be described as an optimization problem, where the aim is to minimize the Kullback-Leibler divergence between $q_\lambda(\bm{\theta})$ and $p(\bm{\theta}|\bm{y})$ with respect to $\bm{\lambda}$, defined as
\begin{align*}
\text{KL}(q_\lambda(\bm{\theta})||p(\bm{\theta}|\bm{y}) ) & = E_{q_\lambda}\left[ \log \frac{q_\lambda(\bm{\theta})}{p(\bm{\theta}|\bm{y})}\right].
\end{align*}
This divergence can be re-written as
$\text{KL}(q_\lambda(\bm{\theta}) ||p(\bm{\theta}|\bm{y}) )  =  \log p(\bm{y})-\calL(\bm{\lambda}) \label{kldexpression}
$,  with $p(\bm{y})=\int p(\bm{\theta})p(\bm{y}|\bm{\theta}) d\bm{\theta}$, $h\left(\bm{\theta}\right)=p\left(\bm{y}|\bm{\theta}\right)p\left(\bm{\theta}\right)$ and where $\calL(\bm{\lambda})$ is known as the Evidence Lower Bound (ELBO), given as
\begin{equation}
\calL(\bm{\lambda})=E_{q_\lambda}\left[\log h(\bm{\theta}) - \log q_\lambda(\bm{\theta})\right]\ .
\label{eq:lowerbound}
\end{equation}
Because $\log p(\bm{y})$ does not depend on $\bm{\lambda}$, minimization of the Kullback-Leibler
divergence with respect to $\bm{\lambda}$ is equivalent to maximizing the ELBO; however, ELBO optimization is more computationally feasible as it does not require evaluation of the intractable term $\log p(\bm{y})$.

Stochastic gradient ascent methods (SGA) can be applied for maximization of the ELBO, by first setting an initial value $\bm{\lambda}^{(0)}$ for $\bm{\lambda}$, and then sequentially iterating over the expression
\begin{align*}
\bm{\lambda}^{(i+1)} & = \bm{\lambda}^{(i)}+\bm{\rho}_i \circ \widehat{\nabla_\lambda \calL(\bm{\lambda}^{(i)})},
\;\mbox{ for } i=1,2,\ldots\,,
\end{align*}
where $\rhovec_i=(\rho_{i1},\dots, \rho_{im})^\top$ denotes a vector of learning rates, `$\circ$' denotes the element-wise product of two vectors, and $\widehat{\nabla_\lambda \calL(\bm{\lambda}^{(i)})}$ is an unbiased estimate of the gradient of $\calL(\bm{\lambda})$ evaluated at $\bm{\lambda}=\bm{\lambda}^{(i)}$.  
Here, the learning rates are set according to the ADAM method proposed by \cite{kingma2014adam}.

The selection of an unbiased and low variance estimate of the ELBO gradient is key to the success of VI. We follow \cite{kingma2013auto}
and employ the so called ``reparametrization trick'' .  
This approach requires a generative formula  $\bm{\theta}=k(\bm{\varepsilon},\bm{\lambda})$ from the specific approximating density $q_\lambda$, where the vector $\bm{\varepsilon}$ has density $f_\varepsilon$ which does not depend on $\bm{\lambda}$. The reparametrization trick then allows to re-write the ELBO as
\begin{align}
\calL(\bm{\lambda}) & = E_{f_\varepsilon}\left[\log h(k(\bm{\varepsilon},\bm{\lambda}))-\log q_\lambda(k(\bm{\varepsilon},\bm{\lambda}))\right]\,. \label{lbdrepar}
\end{align}
By differentiating (\ref{lbdrepar}), the ELBO gradient can be expressed as
\begin{align}
\nabla_\lambda \calL(\bm{\lambda}) & = E_{f_\varepsilon}\left[\frac{\partial\bm{\theta}}{\partial\bm{\lambda}}^\top \left\{\nabla_{\theta} \log h(\bm{\theta})-\nabla_{\theta}\log q_\lambda(\bm{\theta})\right\}\right]\,. \label{lbdgradexpr}
\end{align}
Then, an unbiased low variance estimate of the ELBO gradient can be computed by estimating the expectation in (\ref{lbdgradexpr}) using one random draw from $f_\varepsilon$. 

The last component of VI is the selection of an adequate approximating family.
We follow \cite{ong2018gaussian} and employ a Gaussian approximation with a factor covariance structure, so that
\begin{equation}
q_\lambda(\bm{\theta})=\phi_{m}\left(\bm{\theta};\bm{\mu},BB'+D^2\right)\,,
\label{eq:q}
\end{equation} 
where $\phi_{m}\left(\bm{x};\bm{\mu},\Sigma\right)$, denotes the $m-$variate Gaussian density with mean $\bm{\mu}$ and covariance $\Sigma$, $D$ is a diagonal matrix with diagonal elements $\bm{d} = \left(d_1,\dots,d_m\right)^\top$, $B$ is an $m\times K$ matrix, $K<m$ denotes the number of factors, and $\bm{\lambda} = (\bm{\mu}^\top,\bm{d}^\top,\text{vech}(B)^\top)^\top$, where $\text{vech}$ denotes the half-vectorization operator of a rectangular matrix. With this approximation, the generative formula needed for the reparametrization trick is $\bm{\theta}=k(\bm{\varepsilon},\bm{\lambda}) = \bm{\mu}+B\bm{z}+D\bm{\eta}$, where $\bm{\varepsilon} = \left(\bm{z}^\top,\bm{\eta}^\top\right)^\top$.

The terms needed to perform SGA are $\frac{\partial\bm{\theta}}{\partial\bm{\lambda}}$, $\nabla_{\theta}\log q_\lambda(\bm{\theta})$ and  $\nabla_{\theta} \log h(\bm{\theta})$.
The first two terms were provided in \cite{ong2018gaussian} for the family of approximations used here.
The third term can be re-written as
\begin{align}\label{Eq:logpost_gradient}
\nabla_{\theta} \log h(\bm{\theta}) = \nabla_{\theta} \log p\left(\bm{y}|\bm{\theta}\right) + \nabla_{\theta} \log p\left(\bm{\theta}\right).
\end{align}
We employ uniform priors for $\bm{\theta}$, which implies that $\nabla_{\theta} \log p\left(\bm{\theta}\right)=\bm{0}$. The remaining term, $\nabla_{\theta} \log p\left(\bm{y}|\bm{\theta}\right)$, can be computed as
\begin{align*}
\nabla_\theta\log p\left(\bm{y}|\bm{\theta}\right)
= &\sum_{\left\{i:y_i=1\right\}}\nabla_\theta f_{NN}\left(\bm{x}_i,\bm{\theta}\right)\\
&-\sum_{i}g\left[f_{NN}\left(\bm{x}_i,\bm{\theta}\right)\right]\nabla_\theta f_{NN}\left(\bm{x}_i,\bm{\theta}\right),
\end{align*}
where $\nabla_\theta f_{NN}\left(\bm{x}_i,\bm{\theta}\right)$ is the gradient of the neural network function, which can be evaluated using any readily available back propagation algorithm. 
\subsection{Toy example}
To provide some intuition about how the BNN in Section~\ref{sec:model} can be used in practice, we applied it to a simple simulated data set. 
We started by generating $10000$ realizations of the i.i.d covariates $x_{i,1}\sim N(0,1)$ and $x_{i,2}\sim N(0,1)$.
Subsequently, we generated the corresponding $10000$ realizations of $y_i$ from the true data generating process (DGP) 
\begin{equation}\label{Eq:trueDGP}
    y_i = I(x_{i,1}<t(x_{i,2})+\epsilon_i),
\end{equation}
where $\epsilon_i\sim N(0,1)$ is an i.i.d Gaussian error, while the function $t()$ is the Yeo-Johnson transformation \citep{yeo2000new} with parameter $-1$, whose role is to induce non-linearity in the way the two covariates determine $y_{i}$. 
In order to learn the true DGP in \eqref{Eq:trueDGP}, we applied the BNN to a random subset of $8000$ observations, and used the remaining $2000$ observations to produce out-of-sample predictions. For the choice of $f_{NN}\left(\bm{x}_i,\bm{\theta}\right)$ we used a feedforward neural network with two nodes and three layers.

\begin{figure}\
  \centering
    \includegraphics[scale=0.45]{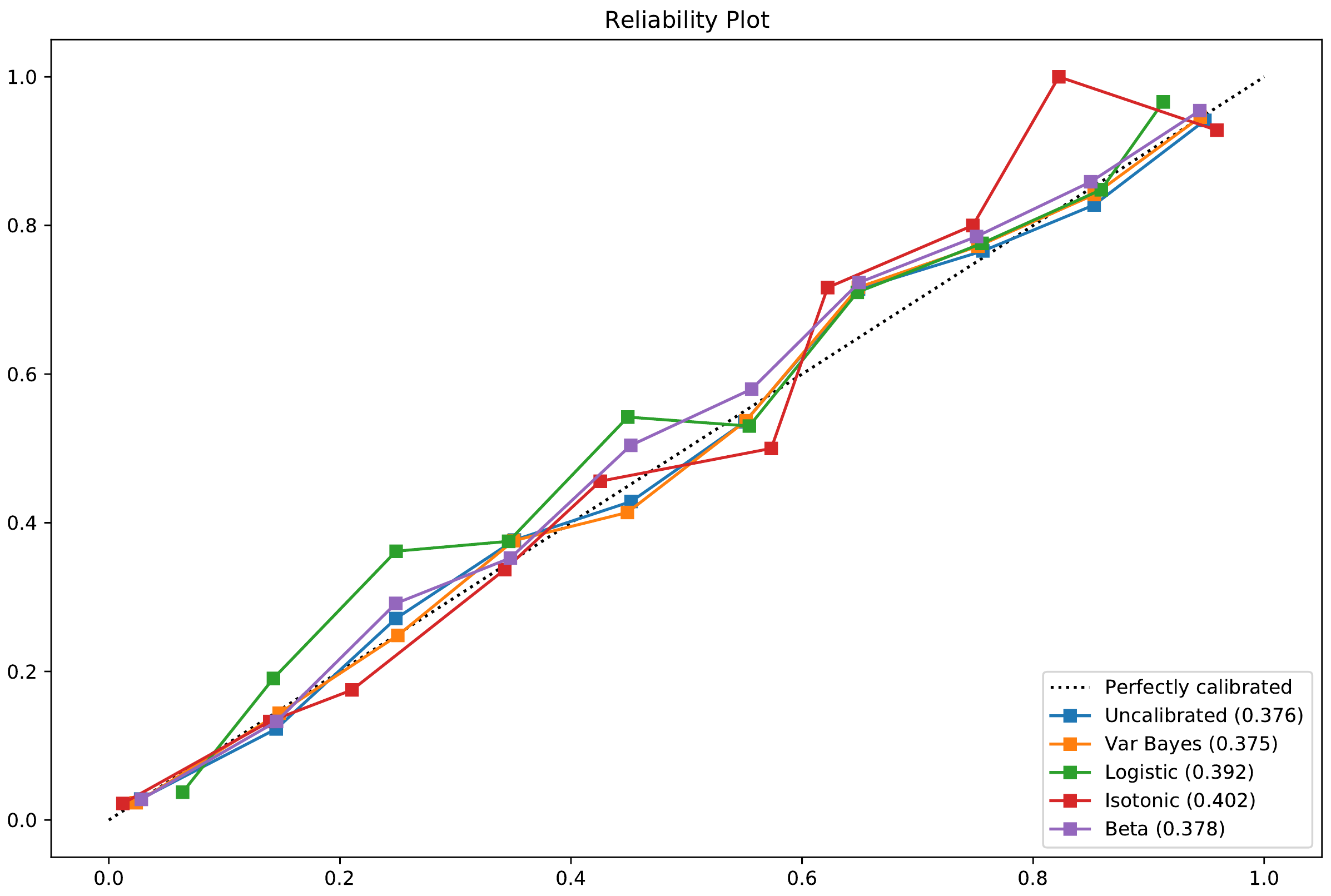}
  \caption{Reliability diagram for the Toy dataset. Also known as Calibration curves, they help visually determine the degree of calibration in probabilistic forecast between different models. Here the x-axis represents the predicted probability and y-axis represents the observed relative frequency of membership to class \cite{brocker2007increasing}. The closer the curves are to perfectly calibrated line (diagonal) the better. The log-losses are denoted in the legend beside each method. We observe here that BNN performs better than the other methods.}.
  \label{fig:reliability-diagram}
\end{figure}

Panel (a) in Figure~\ref{Fig:toyexample} presents the true binary categories of the out-of-sample points as a function of the covariates. The red dots indicate the observations for which $y_i = 1$, while the blue dots display the observations for which $y_i=0$. Although there is some overlap in the classification regions, the red points are concentrated in the top-left quadrant while the blue dots are concentrated in the bottom-right. Panel (b) in Figure~\ref{Fig:toyexample} presents the point predictions $\hat{y}_i$ from the neural network. The point predictions indicate that there is a clear separation threshold. However, we know from panel (a) that the closer the observations are to the threshold, the more classification overlap will be observed and thus the higher the classification uncertainty. Via the computation of the probabilities $\text{Pr}(Y_i=1|\bm{x}_i,\bm{\theta})$, the BNN naturally allow us to measure the classification uncertainty. These probabilities are presented in panel (c) in Figure~\ref{Fig:toyexample}, where the color scale of the dots indicates the classification uncertainty, with green dots indicating higher uncertainty ($\text{Pr}(Y_i=1|\bm{x}_i,\bm{\theta}) \approx 0.5$).
The plot indicates that observations located in the top-left corner are classified as $\hat{y}_i=1$ with high certainty as $\text{Pr}(Y_i=1|\bm{x}_i,\bm{\theta}) \approx 1$. Similarly, observations located in the bottom-right corner are classified as $\hat{y}_i=0$ with high certainty as $\text{Pr}(Y_i=1|\bm{x}_i,\bm{\theta}) \approx 0$. Finally, as the observations get closer to the threshold in the middle, the classification uncertainty increases, which indicates the BNN is able to identify the high probability of class overlap in that region. Figure~\ref{fig:reliability-diagram} shows the reliability diagram for the same highlighting different calibration methods which we discuss in Section~\ref{sec:calibration}.

\section{Post-Hoc probability calibration}
\label{sec:calibration}

We draw upon the seminal work by \cite{Kull2017} to compare our findings with the recent advances in the literature and follow their benchmark methodology.  We benchmark our approach to well established calibration procedures in the literature. The goal of any calibration method is given the output score $s$ of a classifier, to correctly represent the empirical probabilities of a given example belonging to a class. These empirical probabilities can be visualised using reliability diagrams shown in figure \ref{fig:reliability-diagram}. For a given score range from $0.9-1$ on the test set,  90 to 100 percent of the examples should be members of the class.  

\paragraph{Isotonic calibration} is a non parametric method that uses the ranking of the output of any classifier to assign bins for mapping from score to probability based on how well a the classifier has ranked the examples \cite{10.1145/775047.775151,fawcett2007pav}. This approach requires enough samples for any given bin, which for small datasets causes this procedure to overfit\cite{Kull2017}. 

\paragraph{Logistic calibration} was introduced by \cite{platt2000probabilities} for support vector machines. It takes the form of $ \mu(s,\gamma,\delta)=\frac{1}{1+1/exp(\gamma \cdot s + \delta)}$, where $\gamma$ and $\delta$ are real valued parameters. This procedure is commonly referred to as Platt scaling. \cite{Kull2017} showed that this procedure has the tendency to lead to worse results after calibration, compared to only relying on the raw output of a given classifier.

\paragraph{Beta calibration} was introduced by \cite{Kull2017}, where the underlying modeling assumption is the beta distribution, a probability distribution bound to the interval $[0,1]$. In general, the beta distribution is used for modeling the behaviour of proportions or percentages, and therefore suited to model the score distribution for a given class.

Negative log-likelihood (NLL), also referred to as cross-entropy or log-loss, gives us a measure of how well the methods compare. It penalizes predictions that appear to be more confident than they are. In line with the literature \cite{Kull2017}, we use log-loss as our key metric. A measure of accuracy would require the setting of a threshold and might therefore mask the true probabilities of predictions. 

\section{Experiments}
In order to minimize inductive biases imparted by any complex architectural choice, we consider a simple feedforward neural network consisting of just $2$ hidden layers with $4$ ReLU units each. We use ADAM \cite{kingma2014adam} for parameter optimization in both the networks with a global learning rate of $10^{-3}$ or $10^{-2}$ and the following hyperparameters for the same: $\beta_{1} = 0.9$, $\beta_{2} = 0.999$ and $\epsilon = 10^{-7}$. Minimal hyperparameter optimization was done as we mostly selected the default values\footnote{The computation was performed under Ubuntu 18.04 using a Intel(R) Xeon(R) W-2145 CPU @ 3.70GHz and a NVIDIA Quadro P6000 as GPU}.

We devise the following strategy to train both the networks: First, we randomly split the dataset between training ($80\%$) and test ($20\%$) followed by further splitting the train set into validation and calibration sets respectively. For each set, we perform stratified sampling to preserve the original class distribution and standardize the feature matrix with the mean and standard deviation of the train set (avoiding data leakage). During the validation stage, we monitor the cross entropy loss in order to find the optimal number of epochs ($\tau$), which correspond to the minimum validation loss, for each model. Thereafter, we train both the models to $\tau$ and use the base neural network to further calibrate its output probabilities with the post-hoc methods as per Section~\ref{sec:calibration}. We also observed from our experiments that on an average, the standard feedforward neural network needed relatively fewer epochs to converge when compared to the BNN and set their maximum number of epochs during the validation stage to be approximately half that of BNN. We note that early stopping as a regularizing strategy proved to be difficult as the loss curves (particularly for BNNs) sometimes exhibit the double descent phenomena \cite{nakkiran2019deep} to varying degrees.

Furthermore, we employ a similar strategy as \cite{Kull2017} and binarize the target variable, considering the majority class as $1$ and the rest to be $0$, in order to convert from a multi-class to binary classification setting. We also note that calibrated neural networks are exposed to roughly $\sim20\%$ more data points than BNN due to their use of calibration set. This also illustrates that BNNs can still work remarkably well given fewer data points than its counterparts. 

We implement our code using TensorFlow with Python 3.6 to conduct the experiments. In particular, we make use of the fast tensor computation for the Jacobian of errors accessed via the gradient tape on GPU.

\subsection{Datasets}

We evaluate the calibration of all the models on $20$ well-known datasets from the UCI Machine Learning Repository \cite{Dua2019}. Here we consider datasets representing domains such as financial, medical, social, just to name a few, as listed under Table~\ref{uci-datasets-table}.

\begin{table}
  \caption{UCI datasets for used in the experiments.}
  \label{uci-datasets-table}
  \centering
  \begin{tabular}{lrrll}
    \toprule
    Name     & Features     & Samples & Attribute Type(s) & Domain        \\
    \midrule
    Abalone & 8  & 4177 & Categorical, Integer, Real & Life (Marine)  \\
    Balance Scale & 4  & 625 & Categorical & Social  \\
    Credit Approval     & 15 & 653 & Categorical, Integer, Real &  Financial     \\ 
    German Credit  & 20 & 1000 & Categorical, Integer &  Financial     \\ 
    Ionosphere  & 34 & 351 & Integer, Real &  Physical     \\ 
    Image Segmentation & 19 & 2310 & Real &  N/A     \\ 
    Landsat Satellite &  36 & 6435 & Integer & Physical \\
    Letter Recognition &  16 & 35000 & Integer & Computer \\
    Mfeat (Karhunen) &  64 & 2000 & Integer, Real & Computer \\
    Mfeat (Morphological) &  6 & 2000 & Integer, Real & Computer \\
    Mfeat (Zernike) &  47 & 2000 & Integer, Real & Computer \\
    Mushroom &  22 & 8124 & Categorical & Life \\
    Optical Digits Recognition &  64 & 5620 & Integer & Computer \\
    Page Blocks &  10 & 5473 & Integer, Real & Computer \\
    Spambase     & 57 & 4601 & Integer, Real &  Computer     \\
    Vehicle     &   18     & 946   & Integer & N/A \\
    Waveform-5000     &  40     & 5000 & Real & Physical   \\
    WDBC     &  30    &  569 & Real  & Life (Cancer)   \\
    WPBC     &    33    & 194  & Real & Life (Cancer)   \\
    Yeast     &  8      & 1484 & Real  & Life (Proteins)  \\
    \bottomrule
  \end{tabular}
\end{table}

Neural networks typically require training samples orders of magnitude more than those of the UCI datasets. Although theoretically, small sample sizes makes the network prone to overfitting, we empirically did not observe any drastic effect on their performances in part due to the simple architecture with a relatively small number of parameters. We also note that due to the stochastic nature of our learning algorithms they are sensitive to initialization of the parameters.

\subsection{Evaluation}

\begin{figure}
  \centering
    \includegraphics[width=\textwidth]{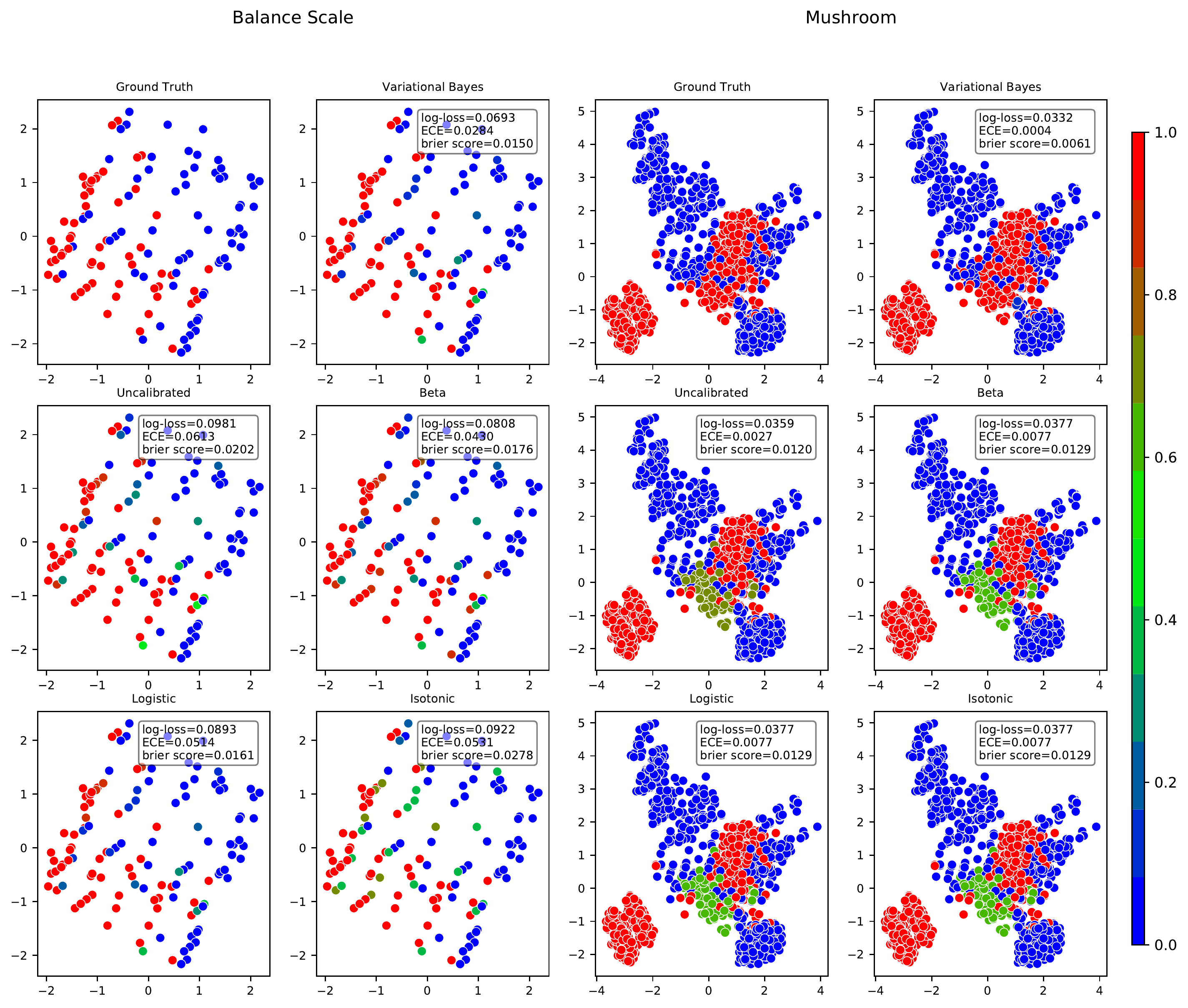}
  \caption{Predictions on the Test set for Balance Scale and Mushroom Datasets. The colour scale represents probabilities between $0-1$. Here we perform PCA on the original feature space and plot on the two principle axes.}
   \label{Fig:predictions}
\end{figure}

We observe that BNN yields competitive performance (and on average outperforms) when compared to the other calibration methods across the datasets as outlined in Table~\ref{tab:results}. As expected, BNN also performs better than the uncalibrated vanilla neural network in most cases. Additionally, we also track other metrics such as Brier score and Expected Calibration Error (ECE) with a bin size of $10$. Brier score provides the mean squared error for a probabilistic forecast and is given by $\frac{1}{N}\sum_{i=1}^{N}(f_i - o_i)$, where $f_i$ are probabilities and $o_i$ are observed classes. ECE is a popular binning based approach to measuring calibration error which is quite sensitive to the choice of bins and ultimately not as strongly reliable \cite{naeini2015obtaining}. Figure~\ref{Fig:predictions} illustrates the performance of each model on two of the datasets. Interestingly, we also notice for a few datasets that calibrating with the post-hoc methods sometimes leads to further miscalibration and higher log-loss.

\begin{table}
   \caption{Log-loss for each dataset. The best method is marked in bold for each row.}
      \label{tab:results}
  \centering
    \begin{tabular}{lrrrrr}
    \toprule
    {} & \multicolumn{5}{c}{Methods}                   \\
    \cmidrule(r){2-6}
     &      Uncalibrated &      Beta &  Isotonic &   Logistic &  Var Bayes \\
     Dataset        &           &           &           &           &           \\
    \midrule
    Abalone         &  \textbf{0.597890}  &  0.598148 &  0.691808 &  0.602227 &  0.615603 \\
    Balance Scale   &  0.098068 &  0.080771 &  0.092153 &  0.089347 &  \textbf{0.069333} \\
    Credit Approval &  0.491384 &  0.522863 &  0.897678 &  \textbf{0.437511} &  0.457618 \\
    German Credit          &  0.525405 &  \textbf{0.517368} &  0.532982 &  0.517461 &  0.565519 \\
    Ionosphere          &  0.385311 &  0.310725 &  0.334547 &  \textbf{0.308752} &  0.330256 \\
    Image Segmentation         &  0.410117 &  0.410120 &  0.410120 &  0.410120 &  \textbf{0.012053} \\
    Landsat Satellite   &  0.549391 &  0.549391 &  0.549391 &  0.549391 &  \textbf{0.065981} \\
    Letter Recognition          &  0.015198 &  0.015066 &  0.015491 &  0.022324 &  \textbf{0.012077} \\
    Mfeat (Karhunen)      &  0.056259 &  \textbf{0.052115} &  0.165397 &  0.121016 &  0.067857 \\
    Mfeat (Morphological) &  0.325083 &  0.325096 &  0.325096 &  0.325096 &  \textbf{0.000206} \\
    Mfeat (Zernike)       &  0.092498 &  0.070683 &  0.166497 &  \textbf{0.048465} &  0.073144 \\
    Mushroom        &  0.035914 &  0.037741 &  0.037741 &  0.037741 &  \textbf{0.033184} \\
    Optical Digits Recognition           &  0.086171 &  0.056051 &  0.086846 &  0.063662 &  \textbf{0.047854} \\
    Page Blocks         &  \textbf{0.071376} &  0.072884 &  0.091231 &  0.082372 &  0.085377 \\
    Spambase        &  0.225907 &  \textbf{0.211673} &  0.212747 &  0.213790 &  0.235523 \\
    Toy             &  0.376422 &  0.377564 &  0.401984 &  0.391937 &  \textbf{0.375313} \\
    Vehicle         &  0.107625 &  0.331832 &  0.310363 &  0.390933 &  \textbf{0.091739} \\
    Waveform-5000   &  0.239015 &  \textbf{0.234735} &  0.240671 &  0.265519 &  0.238321 \\
    WDBC            &  0.080406 &  0.164898 &  0.156011 &  0.176233 &  \textbf{0.070994} \\
    WPBC            &  0.584846 &  0.614227 &  1.777795 &  0.594910 &  \textbf{0.541680} \\
    Yeast           &  0.594895 &  0.583243 &  0.806812 &  \textbf{0.561925} &  0.569832 \\
    \midrule
    Rank           & 2.7619  & 2.6667 &  4.2857 & 3.1905  &  \textbf{2.0952} \\
    \bottomrule
    \end{tabular}
\end{table}

\begin{figure}
  \centering
    \includegraphics[scale=0.5]{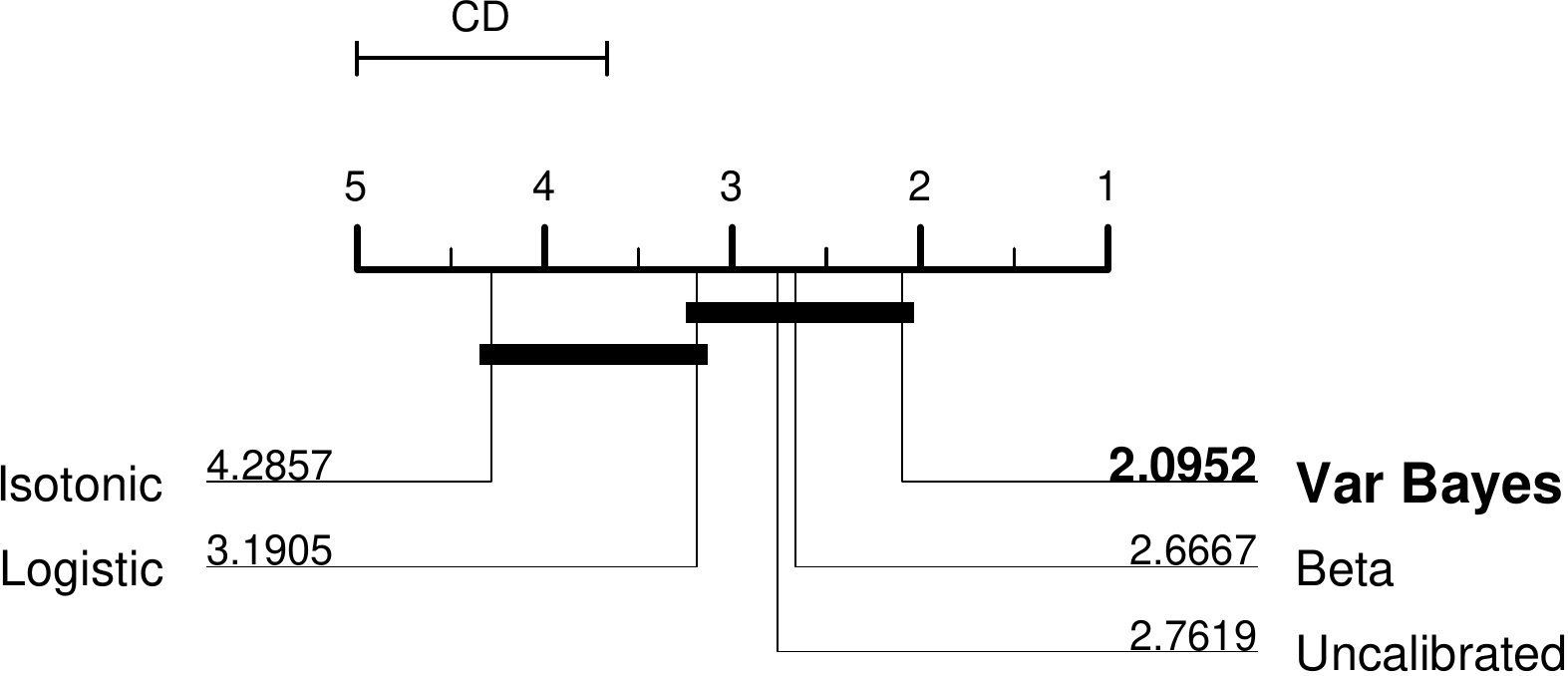}
  \caption{Critical difference diagram for log-loss along with ranks for each method.}
  \label{Fig:critical-diff}
\end{figure}

To establish whether the differences between the methods are statistically significant, we follow \cite{demvsar2006statistical} and perform Friedman test across all datasets with the null hypothesis ($H_0$) as there being no significant differences in performance between the classifiers. Considering a significance level of $0.05$, we reject the null with a p-value of $0.000119$ along with a test statistic of $23.13$ and perform a post-hoc analysis based on Wilcoxon-Holm test to further analyse their pairwise differences. Figure~\ref{Fig:critical-diff} illustrates the critical difference diagram with pairwise significance. Here we observe that BNN (Var Bayes) have the highest rank as compared to the other methods. However, we also note that pairwise differences between BNN, Uncalibrated, Beta and Logistic are not as significant as that between BNN and Isotonic. Overall BNNs provide competitive, if not better, performance as its calibrated counterparts. Interestingly, what \cite{Kull2017} observed for Beta calibration when applied to Adaboost and Naive Bayes classifiers does not perfectly hold for neural nets as evident from our experiments.

\section{Conclusion}

In this paper, we propose the use of BNNs as a plausible alternative to obtain well-calibrated probabilities when compared with post-hoc calibration methods such as platt scaling, isotonic and beta calibration for tabular datasets.

We performed extensive experiments on 20 datasets using the same neural network architecture as our underlying model. With the the normal neural network implementation we applied the common calibration techniques. Our results show that our proposed method works provably well for tabular datasets with small sample size. We made use of recent advances in fast gradient calculation within TensorFlow framework to calculate the full Jacobian matrix with respect to the parameters, as required for Variational Bayes Inference. Albeit, certain datasets required a larger number of epochs for convergence in the Variational framework. Future research will investigate the impact of different priors such as Horseshoe on the parameters and its impact on convergence and calibration.

\section{Ethical and societal implication}
BNNs could be applied across many domains where neural networks are starting to be heavily utilized in real world settings and strict $0$ or $1$ outcomes are not desired. Although for most machine learning researchers it is clear that output scores obtained via softmax, for example, do not represent true probabilities, many practitioners interpret the same as otherwise. This becomes especially critical, when decision threshold are set, that have major consequences for an individual. In areas such as, bail or no bail, treatment with some medicine or not, having wrongly calibrated models  could lead to detrimental or harmful outcomes \cite{rudin2019stop}. Thresholding is also used to try to make machine learning predictions more fair, by setting different thresholds based on protected attributes such as race or gender \cite{hardt2016equality}. Again in such applications, it is highly important that the thresholds represent actual probabilities rather than a ranking. Similarly, in recent years governments have started to scarcely assign resources to its citizen based on algorithmic decisions to make the most efficient use of their limited resources. Hence it becomes increasingly important that resource allocation prioritizes those who truly need such aid.

Alternatively, having well calibrated neural networks could also lead to overconfidence and over reliance on algorithms by decision and policy makers. We would encourage further work of the relationship of calibrated threshold in real world scenarios to fully understand the impact miscalibration has. Nevertheless, we hope our work helps counter these issues and ultimately creates a positive social impact across domains.

\bibliography{main}





\end{document}